\documentclass[letterpaper, 10 pt, conference]{ieeeconf} 
\IEEEoverridecommandlockouts     
\let\labelindent\relax

\usepackage{amsthm}
\usepackage{stmaryrd}
\usepackage{amsmath}
\usepackage{amssymb}
\usepackage{siunitx}

\usepackage{graphicx}
\usepackage{caption}
\usepackage{subcaption}
\usepackage[table]{xcolor}

\usepackage{multirow}

\usepackage{lipsum}  
\usepackage{booktabs}
\usepackage{enumitem}
\usepackage{color}
\usepackage{algorithm}
\usepackage{algorithmic}
\usepackage{pdfpages}

\setlist[itemize]{leftmargin=*}

\usepackage{capt-of}%

\newcommand{\subparagraph}{}

\usepackage{titlesec}
\titlespacing*{\section}{0pt}{0.3\baselineskip}{0.3\baselineskip}
\titlespacing*{\subsection}{0pt}{0.3\baselineskip}{0.3\baselineskip}

\newcommand{\va}{{\bf a}}

\newcommand{\vx}{{\bf x}}
\newcommand{\vy}{{\bf y}}

\newcommand{\vtheta}{\boldsymbol{\theta}}

\title{\LARGE \bf
GraphDistNet: A Graph-based Collision-distance Estimator for \\
Gradient-based Trajectory Optimization
}

\author{Yeseung Kim, Jinwoo Kim, and Daehyung Park\textsuperscript{\textdagger}
\thanks{All authors are with the School of Computing, Korea Advanced Institute of Science and Technology, Korea ({\tt\small \{rhgkrsus1, kjwoo31, daehyung\}@kaist.ac.kr}). {\textsuperscript{\textdagger}}D. Park is the corresponding author.
This work was supported by the National Research Foundation of Korea(NRF) grants funded by the Korea government(MSIT) (No. 2021R1C1C1004368 and 2021R1A4A3032834).}
}

\begin{document}

\maketitle
\thispagestyle{empty}
\pagestyle{empty}

\begin{abstract}
Trajectory optimization (TO) aims to find a sequence of valid states while minimizing costs.
However, its fine validation process is often costly due to computationally expensive collision searches, otherwise coarse searches lower the safety of the system losing a precise solution.
To resolve the issues, we introduce a new collision-distance estimator, GraphDistNet, that can precisely encode the structural information between two geometries by leveraging edge feature-based convolutional operations, and also efficiently predict a batch of collision distances and gradients through $25,000$ random environments with a maximum of $20$ unforeseen objects. Further, we show the adoption of attention mechanism enables our method to be easily generalized in unforeseen complex geometries toward TO. Our evaluation show GraphDistNet outperforms state-of-the-art baseline methods in both simulated and real world tasks.

\end{abstract}
\section{Introduction} \label{sec:intro}

Consider the problem of trajectory optimization (TO) in clutter for manipulation tasks. This requires finding a sequence of feasible configurations with collision avoidance. However, the complexity of the real world reveals the necessity of more precise collision detection, though its computational complexity may lower the usability of the robot. To resolve it, we need a highly-efficient collision detection method that is also robust to diverse environments toward planning of safe and optimal manipulation trajectories.

Traditional detection approaches, such as Gilbert-Johnson-Keerthi (GJK) algorithm \cite{gilbert1988fast}, iteratively find the minimum distance between two sets of sub geometries~\cite{pan2012fcl}. %
There were also efforts to narrow down the search space~\cite{cohen1995collide,bialkowski2016efficient}, sample collidable objects~\cite{kumar2019learning,das2020stochastic,johnson2021chance}, or 
simplify the geometry~\cite{liu2007ellipsoidal} to minimize the iterations. However, simplifications often degrade the completeness or the preciseness of a planning solution. Alternatively, researchers have adopted data-driven models that approximate the detector by modeling the geometric proximity between two point sets~\cite{pan2016fast,pan2015efficient,chase2020neural}. Researchers have also extended the models by adopting incremental/active learning strategies \cite{das2017fastron,das2020learning,zhi2022diffco}. The model approximation becomes more rapid at predicting collisions without exhaustive searches, though the approaches assume a specific environment. Otherwise, they require additional explorations.

\begin{figure}[t]
\centering
\includegraphics[width=0.85\columnwidth]{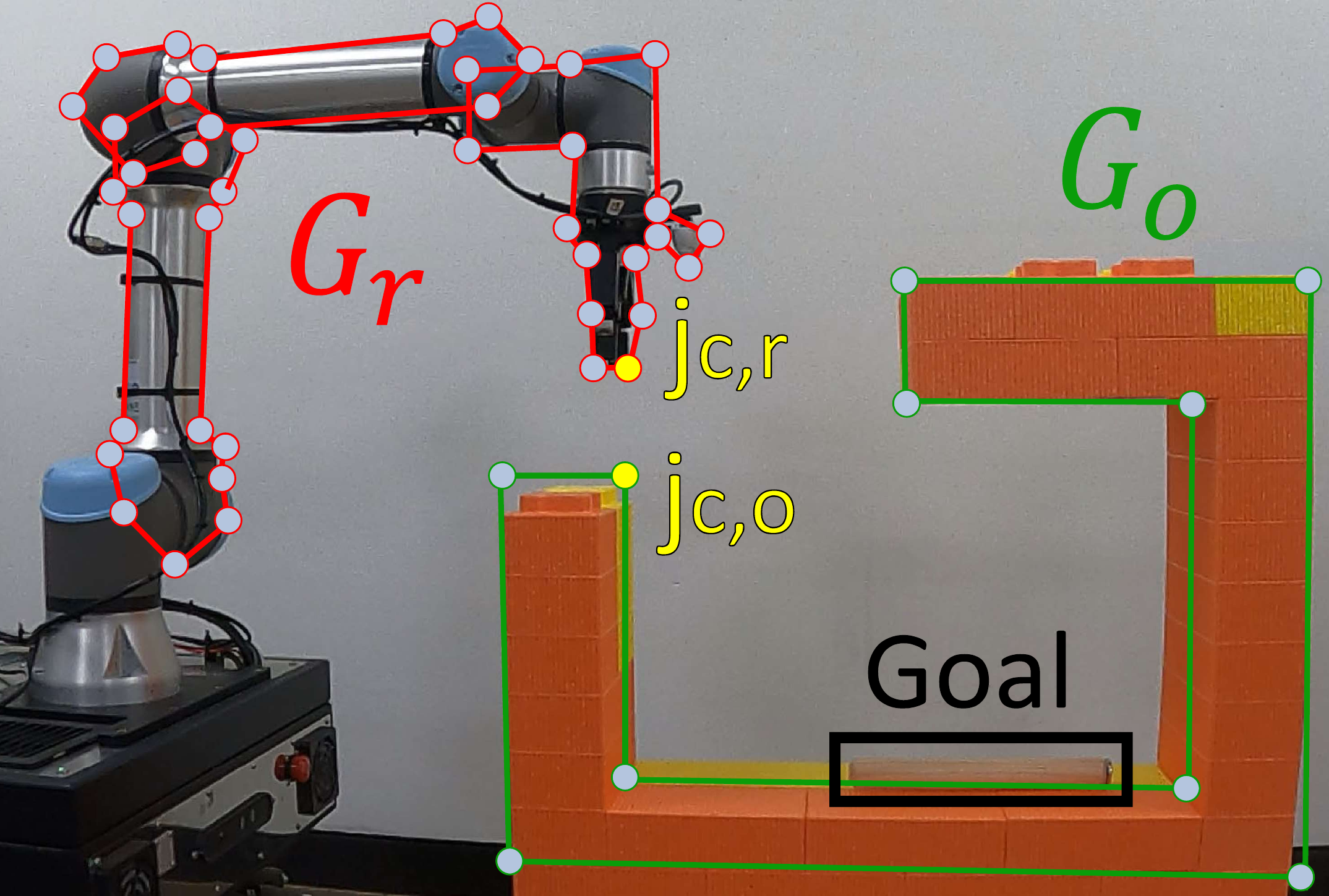}
\caption{\small{An example of the object fetch task. We represent a UR5e manipulator and obstacle geometries as graphs $G_r$ and $G_o$, respectively. Our graph-based collision-distance estimator, GraphDistNet, enables a trajectory optimization method to find collision-free motion by precisely estimating the collision distance and gradients.
}}
\label{fig:representative}
\vspace{-18pt}
\end{figure}

A detector in TO has to be computationally efficient and informative enough to guide the optimization process of trajectory planning. However, most detectors do not provide collision gradients, where the lack of analytical gradients often results in the use of unstable numerical gradients~\cite{schulman2014motion} or stochastic gradients~\cite{stomp2011}. Otherwise, we can use inefficient gradient-free optimization methods. Recently, Zhi et al. introduced the first fully-differentiable detector that finds collision gradients from the distribution of collision scores~\cite{zhi2022diffco}. Although the approach could improve the optimization, the gradient modeling is restricted to priory known geometries modeling the gradients in configuration space.

In this work, we propose a new data-driven collision estimator, GraphDistNet, that precisely estimates the collision distance and gradients between two graphs (e.g., graphs of robot and obstacle geometries). GraphDistNet is the first graph-based collision-distance estimator that can encode the structural relation (i.e., edge feature) between point measures. By constructing a connected robot-object graph, GraphDistNet can embed the geometric relations between graphs via message passing (i.e., convolutional operation) of edge features. In addition, to generalize GraphDistNet to a new environment, we introduce an attention-based graph construction scheme that drastically reduces the construction time and helps to handle more complex geometries without retraining. Finally, we enable GraphDistNet to provide collision gradients to guide TO by leveraging an auto-differentiation tool from deep learning. 

We evaluate our approach on a 2-dimensional (2D) simulated manipulation environment. Also, we show the TO performance given the graphized high-degree of freedom manipulator and object geometries. The proposed detection method is superior to state-of-the-art baseline methods with respect to preciseness, robustness, and generalizability. We also demonstrate the detection and the improved planning performance using a mobile manipulator, \textit{Haetae}, where we show our method enables the robot to safely and also quickly reach a goal in clutter.

Our main contributions are four folds:
\begin{itemize}
    \item We present the first graph-based collision-distance estimation network, GraphDistNet, that precisely regresses the collision distance between objects. %
    \item Our GraphDistNet provides accurate gradients and batch computation results improving TO's performance.
    \item Our GraphDistNet is also robust to various environmental changes and unforeseen environments.
    \item We demonstrate our method outperforms search and learning-based state-of-the-art baseline methods in terms of computational efficiency and preciseness via 2D simulations. We also conduct a proof concept of robotic manipulation tasks in clutter using a real robot. 
\end{itemize}

\section{Related Work}
\label{sec:related}

Trajectory optimization primarily focuses on avoiding obstacles by defining collision penalty or collision-free inequality constraint terms. To specify collision terms, researchers have introduced three representation types of geometric detection approaches: point-based, voxel-based, and mesh-based methods. The first two approaches are suited to non-parametrically describing complex geometries with raw data but do not hold topological information~\cite{pan2013real}. On the other hand, mesh-based methods explicitly represent the surface of a structure, e.g., triangular meshes~\cite{kockara2007collision}. In this field, Gilbert–Johnson–Keerthi (GJK) \cite{gilbert1988fast} and expanding polytope algorithm (EPA) \cite{van2001proximity} are representative types of convex polytope detectors that iteratively find the minimum distance and the penetration depth between two convex sets, respectively. To handle a wide variety of robotic applications, Pan et al. introduced a unified framework, flexible collision library (FCL), that combines GJK, EPA, and other collision detection methods \cite{pan2012fcl}. While these approaches have been widely used in this field, the intricate mesh design increases time and space complexities in the real world. Our method also uses a graph representation to hold topological information, but we adopt a data-driven prediction scheme for faster prediction. 

The data-driven detection is now emerging as the method that has effective representation learning and fast prediction capability~\cite{pan2015efficient, pan2016fast, huh2016learning, das2020learning}. Major approaches often perform kinematic-collision detection in a predetermined environment by modeling collision-free configurations. For modeling, researchers leveraged a wide variety of classic learning approaches: support vector machine~\cite{pan2015efficient, pan2016fast}, Gaussian mixture model~\cite{huh2016learning}, k-nearest neighbor~\cite{pan2016fast}, kernel perceptron~\cite{das2017fastron, das2020learning, zhi2022diffco}, or CMA-ES~\cite{kapusta2019task}. For generalization in new environments, researchers have also adopted incremental/adaptive learning strategies but required additional data collection~\cite{das2017fastron, zhi2022diffco}. Alternatively, Kew et al. introduced a neural network-based collision checker, ClearanceNet, that predicts a collision distance given robot and object configurations~\cite{chase2020neural}. However, ClearanceNet still suffers from varying numbers or shapes of objects. Our method overcomes the generalization issue by looking at a region of interest from a connected graph of robot and object geometries.

The learned model-based detectors can be used for validating candidate trajectories during TO. Major trajectory optimization methods often rely on the quality of collision gradients (i.e., smoothness and continuity) for numerical optimization~\cite{zucker2013chomp, schulman2014motion}. Due to the lack of analytical collision gradients, a major TO method, CHOMP~\cite{schulman2014motion}, precomputes a simplified version of gradients by fitting objects with simpler geometries such as spheres. Similarly, Trajopt~\cite{schulman2014motion} calculates numerical gradients between the swept-volumes of convex or non-convex shapes. However, the computation or the approximation of numerical gradients are costly and unstable, so Trajopt often leads to a local minimum or longer optimization time. Recently, the development of auto-differentiation tools provides a new opportunity by adopting differentiable collision detection. DiffCo~\cite{zhi2022diffco} models a robot's collision and collision-free configuration spaces by adopting a forward-kinematics kernel perceptron. Then, DiffCo generates collision gradients for TO by differentiating its collision-score function. In addition, the batch prediction and gradient computation of ClearanceNet resulted in the acceleration of sampling-based motion planning~\cite{chase2020neural}. We also use neural network based gradient computation and a batch prediction for thousands of collision checks. Furthermore, we improve preciseness and generalization capabilities by adopting graph-based representation and attention techniques.

\section{Background: Graph Neural Networks}
\label{sec:bg}
We briefly review graph neural network (GNN) to regress graph properties. 
Let consider a graph $G=(\mathcal{V},\mathcal{E},\mathcal{X})$ that is a set of nodes $v_i\in \mathcal{V}$, edges $e_{ij}\in \mathcal{E}$, \textit{node features} $\vx_i \in \mathcal{X}$, where the nodes, known as vertices, represent entities or their point measures in a real world. Edges represent nodes' topological relationships such as connectivity, adjacency, and enclosure. The subscripts, $i$ and $j$, represent the indices of nodes in $G$. Likewise, $e_{ij}$ is the edge between $v_i$ and $v_j$.
A node feature $\vx_i$ is a $d_v$-dimensional feature vector ($ \in \mathbb{R}^{d_v}$) associated with $v_i$ (e.g., a point coordinate in pointcloud). However, the node feature often lacks the topological information defined between two nodes, so we can additionally define \textit{edge features} $\vx_{ij} = h_\theta(\vx_i, \vx_j)$ as described in \cite{wang2019dynamic}, where the function $h_\theta$ is a nonlinear mapping function $h_\theta: \mathbb{R}^{d_v}\times \mathbb{R}^{d_v} \rightarrow \mathbb{R}^{d_e}$, $\theta$ is a set of learnable parameters, and $d_e$ is the dimension of edge feature $\vx_{ij}$.
By associating each node and edge with features and aggregating them, we can predict node-, edge-, or graph-level of properties~\cite{zhou2020graph}. 

\begin{figure}[t]
  \centering
  \includegraphics[width=0.8\columnwidth]{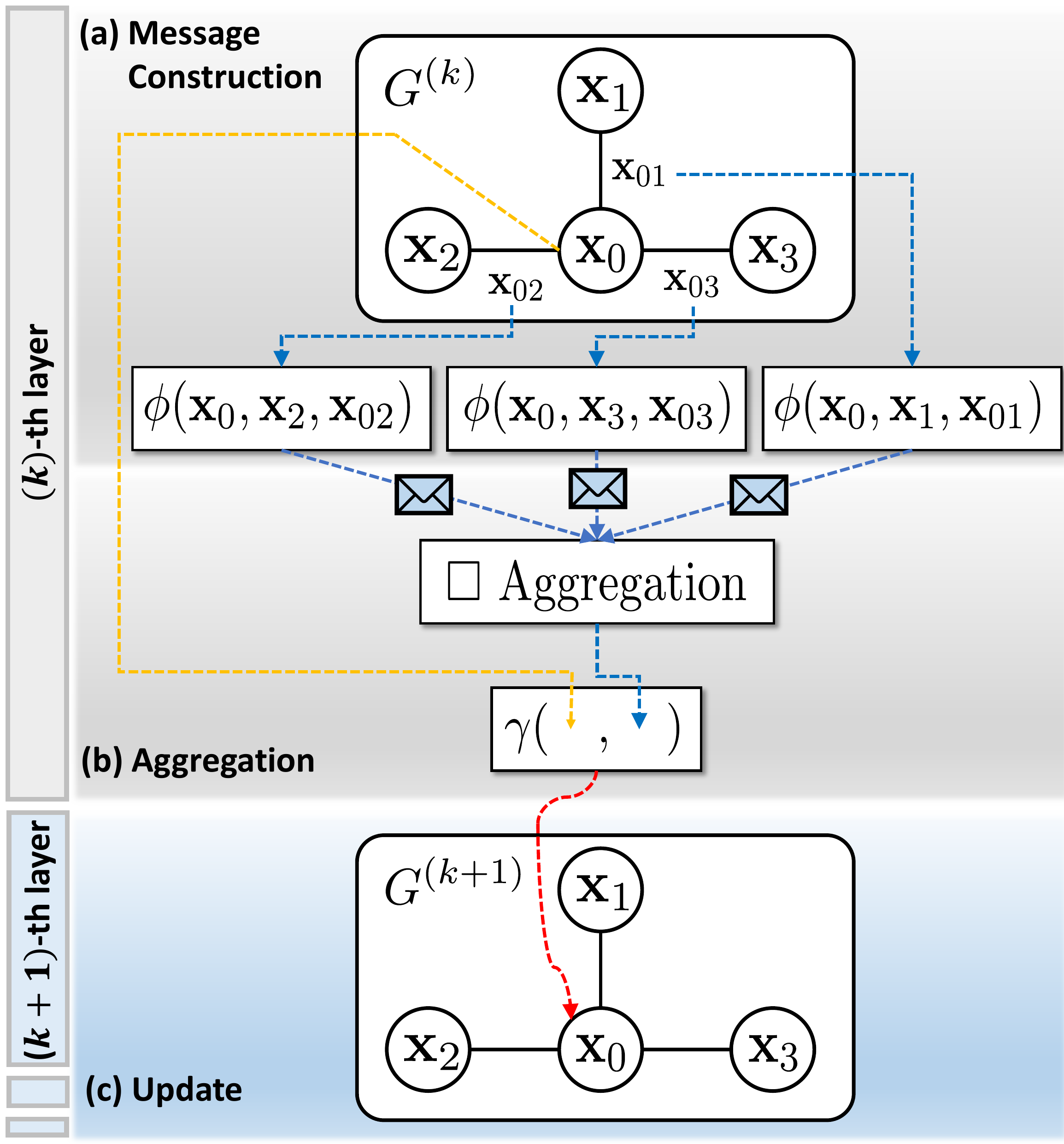}
  \caption{\small{
        A message-passing scheme of convolutional operation at a target node $v_0$ with the node feature $\vx_0$ given a graph $G^{(k)}$. \textbf{(a) Message Construction}: The operation first requires generating a message at each edge around $v_0$ using a function $\phi(\vx_i, \vx_j, \vx_{ij})$. \textbf{(b) Aggregation}: The operation then aggregates the generated messages from (a), and makes an output node feature $\vx_0$ for $G^{(k+1)}$ via a function $\gamma$ with previous feature $\vx_0$. \textbf{(c) Update}: Finally, the operation replaces the target node feature $\vx_0$ with the output of $\gamma$.}
        }
        \label{fig_msg_passing}
        \vspace{-2pt}
\end{figure}

A straightforward aggregation method is \textit{message passing} \cite{gilmer2017neural}, which consists of three steps toward regression/classification (see Fig.~\ref{fig_msg_passing}). First, each node $v_i$ in a graph computes a message (e.g., node or edge features) to exchange it with neighbors. Second, each node aggregates the messages of the neighbors using \textit{sum} or \textit{average} operations. Then, each node updates the node feature using a function of the aggregated message and the current features. We describe a typical form of message passing scheme which is a generalized convolutional operation \cite{fey2019fast} as following: 
\begin{equation}
\mathbf{x}_i^{(k+1)}=\gamma \left( \mathbf{x}_i^{(k)}, \underset{j:(i,j)\in \mathcal{E}}{\square}\phi\left(\mathbf{x}_i^{(k)},\mathbf{x}_j^{(k)},\vx_{ij}^{(k)}\right)\right),
\end{equation}%
where $\gamma$ and $\phi$ are multi-layer perceptrons (MLPs) for the update and aggregation of features, respectively. $\square$ is a symmetric aggregation operator (e.g., $\sum$ or $\max$) and $(k)$ represents the $k$-th layer. Fig.~\ref{fig_msg_passing} shows a visualization of the message passing scheme. The convolutional operation aggregates the messages in the 1-hop neighborhood of a target node $v_j$. Finally, we can collapse the aggregated messages to a vector for regression/classification. Note that $k=0$ means the input layer and $K$ is the number of total layers.

\begin{figure}[t]
\includegraphics[width=0.45\textwidth]{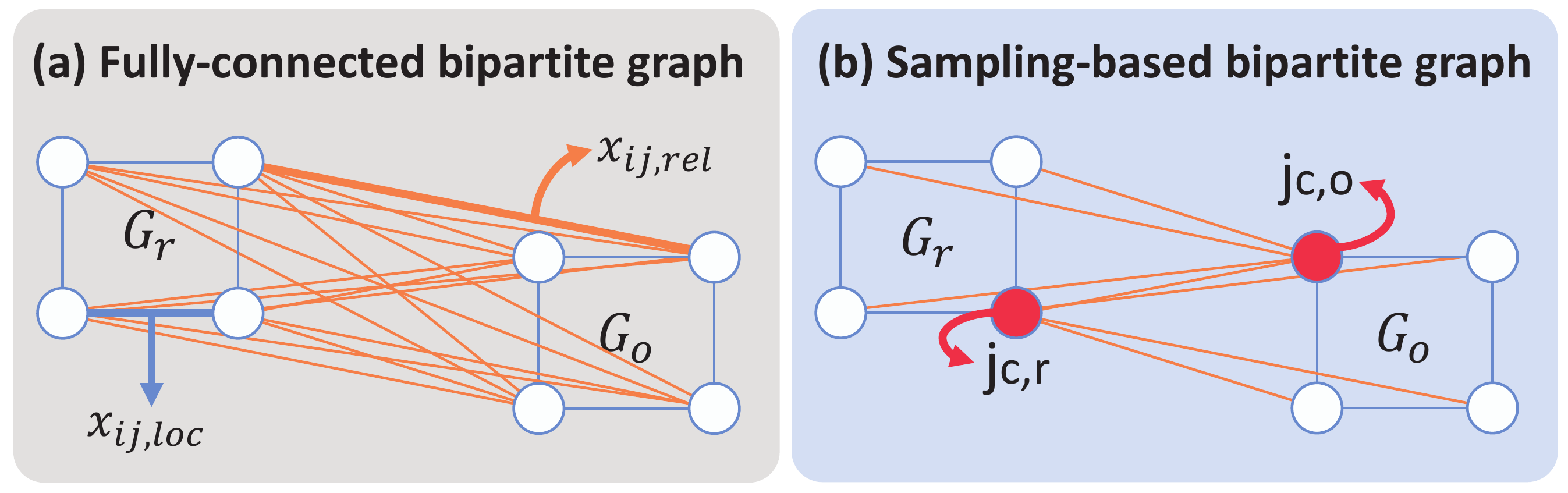} \centering
\caption{\small{Two ways of constructing a connected graph given $G_r$ and $G_o$. \textbf{(a)} We build a fully-connected bipartite graph by creating a maximum number of edges between two graphs. \textbf{(b)} We build a sampling-based bipartite graph by placing a single connection node per graph $(j_{c,r}, j_{c,o})$.
}}
\label{fig:bipartite_graphs}
\vspace{-16pt}
\end{figure}

\begin{figure*}[t]
  \includegraphics[width=\textwidth]{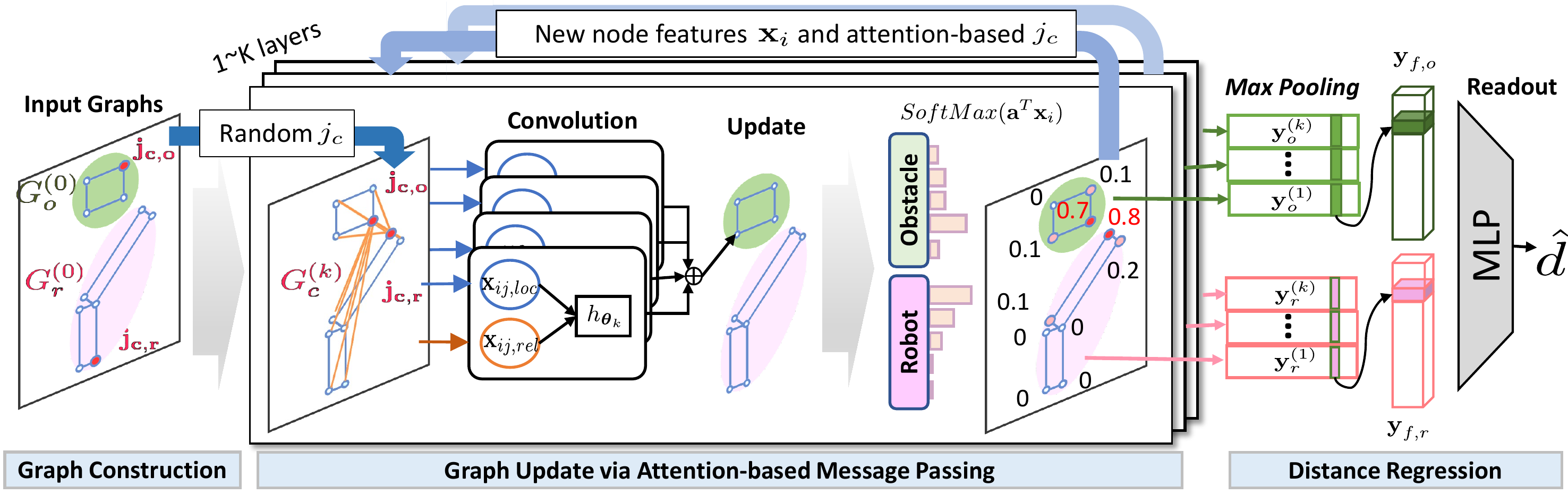} \centering
  \caption{\small{Illustration of a graph-based collision-distance estimation network (GraphDistNet). Given robot and object graphs, our GraphDistNet constructs an initial connected graph $G_u$, and then updates its node features via graph convolutional operations. By selecting the maximally attentive node, which has the highest attention score in each graph, we construct a new connected graph for the convolutional operations on the next layer. Finally, given an iteratively updated graph, our GraphDistNet regresses a collision distance value by aggregating the node features of two graphs and taking the aggregated information into a readout network.  
} }
\label{fig_gdn_layers}
\vspace{-16pt}
\end{figure*}

\section{Method: GraphDistNet}
\label{sec:method}
We introduce a graph-based collision-distance estimator, GraphDistNet, that encodes the geometric relation between two graphs and predicts their minimum distance and collision gradients for TO. In this section, we describe how to 1) construct a connected input graph, 2) encode the geometric relation via message passing, and 3) estimate the collision-distance and gradients toward TO in order. Finally, we discuss the collision distance that can improve the distance regression performance.

\subsection{Initial Graph Construction}
Consider a pair of robot and object geometries that can be represented as undirected graphs, $G_r=(\mathcal{V}_r, \mathcal{E}_r, \mathcal{X}_{r})$ and $G_o=(\mathcal{V}_o, \mathcal{E}_o, \mathcal{X}_{o})$, respectively. Assuming the graphs are independent, $\mathcal{V}_r\cap \mathcal{V}_o=\emptyset$, we construct a connected graph by adding a bipartite graph $G_c$ that connects each node in $G_r$ with all or part of the nodes in $G_o$. We create a connected graph to exchange geometric relationships between two graphs via message passing over edges. 

For the connection, we can first use a fully-connected bipartite graph $G_c$, where each node in geometry connects with all nodes in another geometry (see Fig.~\ref{fig:bipartite_graphs}(a)). 
We can then build a connected graph $G_{u} = G_{r} \cup G_{o} \cup G_c$. However, construction has several drawbacks regarding construction time and space complexities. Given the large graphs due to complex geometries or multiple objects, the space complexity drastically increases to $|\mathcal{E}_c | = |\mathcal{V}_r | \times |\mathcal{V}_o|$, since the fully-connected graph requires connecting all nodes between $\mathcal{V}_r$ and $\mathcal{V}_o$. The enlargement of complexity causes a problem with a high-memory footprint and computationally expensive gather-scatter operations in the message passing \cite{fey2019fast}.

Alternatively, we define a sampling-based bipartite graph as $G_c=(\mathcal{V}_c,\mathcal{E}_c, \mathcal{X}_c)$, where $\mathcal{E}_c$ connects either the nodes of $G_r$ with a node $v_{j_{c,o}}$ of $G_o$ or the nodes of $G_o$ with a node $v_{j_{c,r}}$ of $G_r$ (see Fig.~\ref{fig:bipartite_graphs}(b)):
\begin{equation}
\begin{aligned}
\mathcal{E}_c = \{e_{ij_{c,o}} | &(\forall v_i\in \mathcal{V}_r, \exists v_{j_{c,o}} \in \mathcal{V}_o  )\}\\
&\cup \{e_{ij_{c,r}} |  (\forall v_i \in \mathcal{V}_o, \exists v_{j_{c,r}}\in \mathcal{V}_r) \},
\end{aligned}
\end{equation}
where we randomly select $(j_{c,r}, j_{c,o})$ and $e_{ij}=e_{ji}$ since we use undirected graphs. For the sake of conciseness, we omit the symbol $o$ and $r$ behind so we use $j_c$ to indicate the index of the selected informative node that can be connected to $v_i$.
Note that $v_i$ and $v_{j_c}$ must not be on the same graph.
With a sampling-based bipartite graph, we can still build a connected graph $G_{u}$ (see Fig.~\ref{fig_gdn_layers}). Although we randomly select each $j_c$ for initial graph construction, $j_c$ is not necessarily a random index after the first message passing. In Sec.~\ref{ssec:graph_update}, we will discuss how to construct better message-passing structure by introducing an attention-score based $j_c$ selection mechanism.

To embed geometric information in a message, we define a node feature $\vx_i \in \mathcal{X}_u$ as a Cartesian coordinate $(x_i, y_i) \in \mathbb{R}^2$ of the node $v_i$ in a graph $G_u$, where $\vx_i$ is not necessarily be a 2D coordinate and its dimension varies depending on the convolutional operation. In this work, we gradually update the node features over layers in GraphDistNet. To distinguish each, we define the initial graph as $G_u^{(0)}$ and represent each updated graph as $G_u^{(k)}$, where $k=1, ... , K$.

\subsection{Graph Update via Attention-based Message Passing}\label{ssec:graph_update}
We propose a new message passing method for GraphDistNet that gradually aggregates the information of the minimum distance between $G_r$ and $G_o$ via the $K$ number of message-passing layers. 
The aggregation is a process of convolutional operations, where each node $v_i$ exchanges node/edge features with its 1-hop neighbor nodes. We define the procedure at a target node $v_i$ as following:
\begin{align}
\vx_i^{(k+1)}&=\sum_{j\in \mathcal{N}_{ro}(i)\cup\{i\}} h_{\vtheta_k}^{(k)}\left(\vx_{ij, loc}^{(k)} , \vx_{ij_{c}, rel}^{(k)}\right),
\label{eq_conv_op}
\end{align}
where $\mathcal{N}_{ro}(i)$ is the index set of 1-hop neighborhood of the target node $v_i$ in $G_{ro}$ $(=G_r \cup G_o)$ and $j_c$ is a random or the best node index that satisfies $e_{ij_c}^{(k)} \in \mathcal{E}_c^{(k)}$ and $i \neq j_c$. $\vx_{ij, loc}^{(k)} \in \mathcal{X}_{loc}^{(k)}$ and $\vx_{ij_c, rel}^{(k)} \in \mathcal{X}_{rel}^{(k)}$ are two types of edge features, \textit{local}-edge and \textit{relative}-edge features, respectively. We will describe the details later. The convolution encodes both \textit{local}-edge and \textit{relative}-edge features via an MLP-based encoder $h_{\vtheta_k}^{(k)}$ with a set of trainable parameters $\vtheta_k$ and aggregates the features over the neighborhood of the target node $v_i$. Updating node features per layer, we iteratively obtain new graphs $\left(G_u^{(1)}, G_u^{(2)}, ..., G_u^{(K)} \right)$ in order. Note that the structure of graphs varies since we sample $j_c$ per layer. We describe the details below.

The proposed convolution operation focuses on capturing the local structure among nodes (i.e., edge features), since the distance estimation does not require the information of global structure such as the global coordinates of objects. Assuming each object is a collection of small patches~\cite{wang2019dynamic}, GraphDistNet encodes two types of edge features: \textit{local}-edge feature $\vx_{ij,loc}\in \mathcal{X}_{loc}$ and \textit{relative}-edge feature $\vx_{ij,rel}\in \mathcal{X}_{rel}$ associated with edges in $\mathcal{E}_r \cup \mathcal{E}_o$ and $\mathcal{E}_c$, respectively:
\begin{align}
\vx_{ij,loc}^{(k)} &= \vx_{ij}^{(k)} \text{ if $e_{ij} \in \mathcal{E}_r \cup \mathcal{E}_o$},\\
\vx_{ij,rel}^{(k)} &= \vx_{ij}^{(k)} \text{ if $e_{ij} \in \mathcal{E}_c$}.
\end{align}
Both features represent the difference between two node features and are used for updating previous node features via message passing. To minimize the degradation problem of losing the original geometric information during the iterative message passing, we use an improved version of the edge feature that is a concatenated vector of the current difference and a shortcut connection difference from the initial graph $G_u^{(0)}$: 
\begin{align}
\vx_{ij}^{(k)} =
\begin{cases}
\left( \vx_j^{(0)}-\vx_i^{(0)} \right)  & \text{if $k=0$},\\
\left( \vx_j^{(0)}-\vx_i^{(0)} \mathbin\Vert \vx_j^{(k)}-\vx_i^{(k)} \right)  & \text{otherwise},
\end{cases}
\end{align}
where $j$ represents a node index in 1-hop neighbors of the target node $v_i$, the operator $\mathbin\Vert$ represents the concatenation of two vectors, and $\vx_{ij}^{(0)} \in\mathbb{R}^2$.  

The convolutional operation in Eq.~(\ref{eq_conv_op}) aggregates edge features and updates node features via an MLP-based encoder $h_{\vtheta_k}^{(k)}$ at the $k$-th layer: 
\begin{align}
h_{\vtheta_k}^{(k)}:
\begin{cases}
 \mathbb{R}^4 \rightarrow \mathbb{R}^{d_h}  & \text{if $k=0$},\\
 \mathbb{R}^{2d_h+4} \rightarrow \mathbb{R}^{d_h}  & \text{otherwise},
\end{cases}
\end{align}
where $d_h=32$. In this work, we use a $(32, 32, 32)$ size of 3- hidden layer MLPs with a Leaky ReLU activation function where its slope parameter is $0.2$. 

After each convolution, we find the most informative $v_{j_c}$ in each graph $G_r$ and $G_o$ by introducing an attention-based selection method. The informativeness of \textit{relative}-edge features depends on the selection of $j_c$, which enables GraphDistNet to exchange the collision-distance information between $G_r$ and $G_o$. However, the uniform random selection may degrade the message-passing performance by finding far-distance edges. Further, a large number of obstacles or complex geometries require a large number of layers to pass necessary features. Therefore, we adopt graph attention~\cite{velickovic2018graph} and sampling~\cite{serafini2021scalable} schemes, where the attention mechanism helps finding the most relevant node to predict the collision distance by computing an attention score per node. To do that, our method first computes an attention score per node feature in each graph as $\va^{(k)T} \vx_i^{(k)}$, where $i$ is the node index in $G_{u}^{(k)}$ and $\va^{(k)}$ is a vector of trainable parameters ($\in \mathbb{R}^{d_h}$) in the attention mechanism. By using $\va^{(k)}$, we can determine the attention weights of node features in each graph $G_r^{(k)}$ and $G_o^{(k)}$ as follows:
\begin{align}
\alpha_{i_r}^{(k)}&=\frac{\exp(\va^{(k)T} \vx_{i_r}^{(k)})}{\sum_{j_r}\exp(\va^{(k)T} \vx_{j_r}^{(k)})},\\
\alpha_{i_o}^{(k)}&=\frac{\exp(\va^{(k)T} \vx_{i_o}^{(k)})}{\sum_{j_o}\exp(\va^{(k)T} \vx_{j_o}^{(k)})}, 
\end{align}
where $v_{i \in \{i_r, j_r\}}^{(k)}\in \mathcal{V}_r^{(k)}$ and $v_{i \in \{i_o, j_o\}}^{(k)}\in \mathcal{V}_o^{(k)}$. $\alpha_{i_r}^{(k)}$ and $\alpha_{i_o}^{(k)}$ are attention weights for $G_r^{(k)}$ and $G_o^{(k)}$, respectively. The node index with the maximum attention weight indicates $j_c$ in each graph (e.g., $j_{c,r} = \arg\max_{i_r} \alpha_{i_r} $ and $j_{c,o} = \arg\max_{i_o} \alpha_{i_o} $).

\subsection{Collision-distance and -gradient Estimation}
We regress a collision distance via a readout network by adopting the node features of layerwise output graphs $\left( (G_r^{(0)}, G_o^{(0)}), ..., (G_r^{(K)}, G_o^{(K)}) \right)$. To do that, we compute the attention-weighted features to extract the  informative features from each graph by applying an attention mechanism: 
\begin{align}
\vy_r^{(k)} &= f_{LR} \left( \sum_{i_r}\alpha_{i_r}\vx_{i_r}^{(k)}\right) \in \mathbb{R}^{d_h}, \\
\vy_o^{(k)} &= f_{LR} \left( \sum_{i_o}\alpha_{i_o}\vx_{i_o}^{(k)}\right) \in \mathbb{R}^{d_h}, 
\end{align}
where $\vx_{i_r}^{(k)} \in \mathbb{R}^{d_h}$, $\vx_{i_o}^{(k)} \in \mathbb{R}^{d_h}$, and $f_{LR}$ is a LeakyReLU activation function with a slope parameter $0.2$. After stacking the attention-weighted features $\left( \vy_r^{(1)T}, ..., \vy_r^{(K)T} \right)\in \mathbb{R}^{d_h \times K}$, we perform element-wise max pooling to hold permutation invariance. Note that we use the stacked features to backpropagate the distance error to each layer via skip connections. We then obtain a $d_h$-dimensional vector, $\vy_{f,r}\in \mathbb{R}^{d_h}$. Likewise, we also compute $\vy_{f,o}\in \mathbb{R}^{d_h}$. Finally, we estimate a collision distance,
\begin{align}
\hat{d} = MLP\Big( \underbrace{\bigparallel_{i=1}^{d_h} \max_k \vy_r^{k}(i)}_{\vy_{f,r}} \mathbin\Vert \underbrace{\bigparallel_{i=1}^{d_h} \max_k \vy_o^{k}(i)}_{\vy_{f,o}} \Big),
\end{align}
where we use a $(32,32,32)$ size of 3-hidden layer MLPs. The output $\hat{d}$ indicates an estimated collision distance.

We can then derive additional tools from GraphDistNet that can benefit trajectory planning. One is a binary collision checker setting a safety distance-margin $d_{margin}$:
\begin{align}
\text{Is collision? = }
\begin{cases}
\text{True} & \text{ if }\hat{d} \leq d_{margin}, \nonumber \\
\text{False} & \text{ otherwise},  \nonumber
\end{cases}
\end{align}
where the margin $d_{margin}$ varies depending on the robot and objects. We can say a robot is in a collision when the collision distance $\hat{d}$ is within the margin. Another tool is a collision-gradient estimator that returns a gradient of distance measure with respect to a configuration of the robot. We can compute the gradient using any automatic differentiation tool in deep learning libraries since GraphDistNet is analytically differentiable. Thus, we can obtain a gradient as $\partial GraphDistNet(\vtheta) / \partial \vtheta$ given a joint configuration $\vtheta$ that can build a robot graph $G_r$. Providing the gradients for collision avoidance, gradient-based trajectory optimization solvers can be faster than using conventional numerical gradients such as finite differences.

\subsection{Collision Distance}
The training of our GraphDistNet requires a pair of input graphs and output distance. The input is a set of a robot graph $G_r$ and a union of object graphs, $G_o = G_{o,1} \cup ... \cup G_{o,n} $, where $n$ is the number of objects, and the output can be the minimum distance between $G_r$ and the nearest object $G_{o,i} \subset G_o$. However, the minimum distance is always zero when two target geometries overlap regardless of the degree of penetration. This causes non-smooth regression that degrades the estimation performance near the collision status. To resolve the problem, we use a modified collision-distance function $f_{cd}$ similar to \cite{zhi2022diffco}:
\begin{equation}
f_{cd}(G_r, G_o)=
\begin{cases}
    \min_i f_{d}(G_r, G_{o,i}), & \text{if $G_r \cap G_{o} = \emptyset $},\\
    -\max_i f_{pd}(G_r, G_{o,i}), & \text{otherwise},
\end{cases} \nonumber
\end{equation}
where $f_{d}(\cdot)$ returns the minimum distance between input geometries and $f_{pd}(\cdot)$ returns the local translational-penetration depth computed from flexible collision library (FCL)~ \cite{pan2012fcl}. %

\section{Evaluation Setup} \label{sec:exp_setup}

\newcolumntype{C}[1]{>{\centering\arraybackslash}m{#1}}
\begin{table*}[ht]
\begin{minipage}[c]{0.31\linewidth}
\centering
\includegraphics[width=1.0\columnwidth]{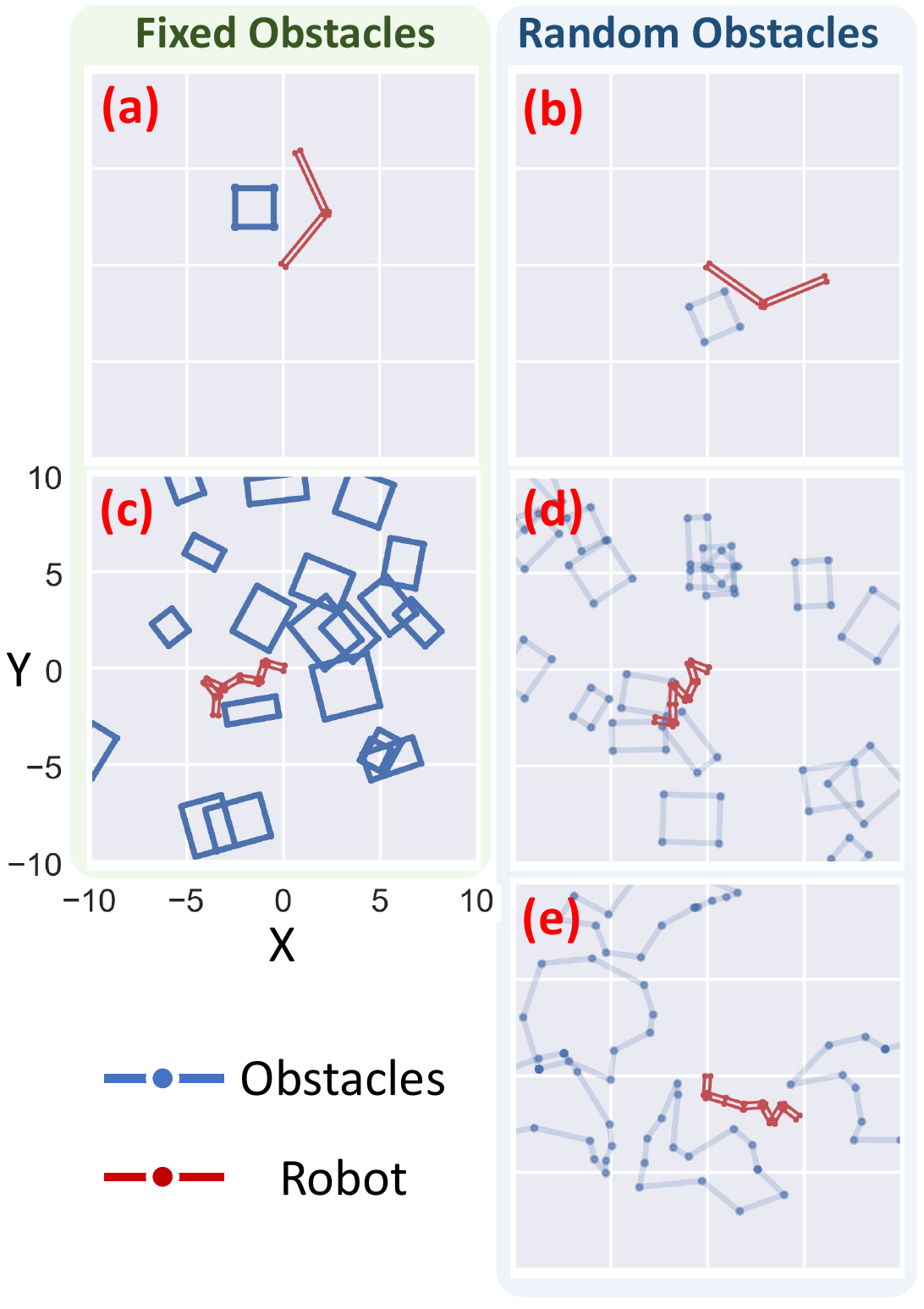}
\captionof{figure}{\small{Examples of 2D planar arm environments. Robots and objects are red and blue, respectively. For (a) and (c), we use the same objects for both training and test settings. For (b), (d), and (e), we vary the pose and shape of objects. }
}
\label{fig_col_dist_est_env}
\end{minipage}
\hfill
\begin{minipage}[c]{0.67\linewidth}
\centering
\footnotesize
  \begin{tabular}{c l C{0.9cm} C{2.5cm} c c c} 
     \toprule
     \multirow{3}{*}{Env.} & \multirow{3}{*}{Method} & \multicolumn{3}{c}{Distance Regression \& Classification} & Gradient Est. \\     \cmidrule(lr){3-5} \cmidrule(lr){6-6} 
      &  & {\scriptsize Elapsed}  & MAE & \multirow{2}{*}{AUC} & \multirow{2}{*}{ACD} \\
      &  & {\scriptsize Time (s)}  & with \textit{p}-value & &  \\      
     \midrule
      \multirow{4}{*}{\shortstack[lb]{Fig.~\ref{fig_col_dist_est_env}\\(a)}} 
                &  FCL              & 1.0653 & - & - & - \\
                & DiffCo           & 0.0048 & \multicolumn{1}{l}{\quad\; N/A} & 1.0000 & 0.3575 \\ %
                & ClearanceNet     & \textbf{0.0009} &  \multirow{2}{*}{$\left.
                \begin{array}{l}
                    0.0082 \\              
                    \textbf{0.0031} 
                \end{array}\right\rbrace\scriptscriptstyle 
                \num{6e-20}
                $} &1.0000  & 0.0255 \\ %
                & GraphDistNet & 0.0115 &  & \textbf{1.0000} & \textbf{0.0054} \\ %
      \hline
      \multirow{5}{*}{\shortstack[lb]{Fig.~\ref{fig_col_dist_est_env}\\(b)}} 
                & FCL & 1.1822 & - & - & - \\
                & DiffCo {\scriptsize (w/o active learning)} & 0.0054 & \multicolumn{1}{l}{\quad\; N/A} & 0.5031 & 1.0270\\ %
                & DiffCo {\scriptsize (w/ active learning)} & 2058.2 & \multicolumn{1}{l}{\quad\; N/A} & 0.9948 & 0.4221\\ %
                & ClearanceNet & \textbf{0.0009} & \multirow{2}{*}{$\left.
                \begin{array}{l}
                    0.1732 \\              
                    \textbf{0.0390} 
                \end{array}\right\rbrace\scriptscriptstyle 
                \num{2e-12}
                $}
                & 0.9986 & 0.2853\\ %
                & GraphDistNet & 0.0115 &  & \textbf{0.9999} & \textbf{0.1748}\\ %
      \hline
      \multirow{4}{*}{\shortstack[lb]{Fig.~\ref{fig_col_dist_est_env}\\(c)}} 
                & FCL & 3.1573 & - & - & - \\
                & DiffCo           & 0.0740 & \multicolumn{1}{l}{\quad\; N/A} & 0.9843 & 0.3202\\ %
                & ClearanceNet     & \textbf{0.0012} & \multirow{2}{*}{$\left.
                \begin{array}{l}
                    0.2360 \\              
                    \textbf{0.0253} 
                \end{array}\right\rbrace\scriptscriptstyle 
                \num{6e-60}
                $} & 0.9111 & 0.5933\\ %
                & GraphDistNet & 0.0574 &  & \textbf{0.9990} & \textbf{0.1458}\\ %
      \hline      
      \multirow{5}{*}{\shortstack[lb]{Fig.~\ref{fig_col_dist_est_env}\\(d)}} 
                & FCL & 4.4467 & - & - & - \\
                & DiffCo {\scriptsize (w/o active learning)}  & 0.0742 & \multicolumn{1}{l}{\quad\; N/A} & 0.5239 & 0.9450\\ %
                & DiffCo {\scriptsize (w/ active learning)}  & 75129.5 & \multicolumn{1}{l}{\quad\; N/A} & 0.9798 & \textbf{0.3968}\\ %
                & ClearanceNet* & \textbf{0.0012} & \multirow{2}{*}{$\left.
                \begin{array}{l}
                    0.7725 \\              
                    \textbf{0.1614} 
                \end{array}\right\rbrace\scriptscriptstyle 
                \num{1e-10}
                $} & 0.5691 & 0.9491\\ %
                & GraphDistNet & 0.0546 &  & \textbf{0.9842} & 0.4152\\ %
      \hline
      \multirow{5}{*}{\shortstack[lb]{Fig.~\ref{fig_col_dist_est_env}\\(e)}} 
                & FCL & 6.8432 & - & - & - \\
                & DiffCo {\scriptsize (w/o active learning)} & 0.0350 & \multicolumn{1}{l}{\quad\; N/A} & 0.5200 & 0.9612\\ %
                & DiffCo {\scriptsize (w/ active learning)} & \multicolumn{4}{c}{\cellcolor{gray!25} \footnotesize{(Failure: Out of Memory)}} \\
                & ClearanceNet* & \textbf{0.0011} & \multirow{2}{*}{$\left.
                \begin{array}{l}
                    0.7479 \\              
                    \textbf{0.2446} 
                \end{array}\right\rbrace\scriptscriptstyle 
                \num{3e-2}
                $} & 0.8882 & 0.9767\\ %
                & GraphDistNet & 0.1251 &  & \textbf{0.9925} & \textbf{0.8421}\\ %
     \bottomrule
  \end{tabular}
  \caption{\small{Comparison of the GraphDistNet and baseline methods on planar arm environments. We evaluated each method with 5000 random samples, where the elapsed time represents the average computation time of the 5000 estimation samples. MAE represents the mean absolute error between the estimated collision distance and its ground truth from FCL. \textit{p}-value represents the statistical significance of error distributions via Welch's \textit{t}-test. AUC is the area under the receiver operating characteristic (ROC) curve metric for the binary collision classification problems. The ACD is the average of cosine distances between the estimated gradient fields and their ground truths from FCL. Note that MAE is not available in DiffCo as DiffCo returns a score instead of a distance. (*we applied dropouts with $p=0.5$.)}}
  \label{table_dist_grad_comp}
\end{minipage}
\vspace{-18pt}
\end{table*}

\subsection{Collision Distance and Gradient Estimation}
We first aimed to evaluate the accuracy and the robustness of the estimated collision distance and gradients via planar arm environments. Fig.~\ref{fig_col_dist_est_env} shows five types of environments where we placed either 2 or 7 degree-of-freedom (DoF) planar arms with obstacles. We measured the distance and gradient estimation performance by randomly placing single or multiple objects. We describe each setting as follows:
\begin{itemize}
\item \textbf{2-DoF Arm \& 1-fixed Obj.}: We placed a 2-DoF planar arm and a fixed square object, as shown in Fig.~\ref{fig_col_dist_est_env}(a). 
\item \textbf{2-DoF Arm \& 1-random Obj.}: We placed a 2-DoF planar arm and a random position and orientation of the square object, as shown in Fig.~\ref{fig_col_dist_est_env}(b).
\item \textbf{7-DoF Arm \& 20-fixed Objs.}: We placed a 7-DoF planar arm and a fixed distribution of 20 square objects, as shown in Fig.~\ref{fig_col_dist_est_env}(c). 
\item \textbf{7-DoF Arm \& 20-random Objs.}: We placed a 7-DoF planar arm and randomly distributed and rotated 20 square objects, as shown in Fig.~\ref{fig_col_dist_est_env}(d)
\item \textbf{7-DoF Arm \& 5 random $\alpha$-shapes}: We placed a 7-DoF planar arm and 5 randomly-created $\alpha$-shape~\cite{edelsbrunner1994three} objects, as shown in Fig.~\ref{fig_col_dist_est_env}(e). To create an $\alpha$-shape, we sampled 20 points uniformly in 2D space and connected maximum 15 points with $\alpha=0.6$ to make a closed loop of points as an object. We then removed the remaining.
\end{itemize}

We trained our method and two learning-based baselines (i.e., DiffCo~\cite{zhi2022diffco} and ClearanceNet~\cite{chase2020neural}) with $50,000$ number of random configurations of the robot and objects. We then tested methods with $5,000$ random configurations of environments. Note that DiffCo requires training in a fixed environment so we trained DiffCo on Fig.~\ref{fig_col_dist_est_env}(a) and Fig.~\ref{fig_col_dist_est_env}(c) for the evaluations on Fig.~\ref{fig_col_dist_est_env}(b) and Fig.~\ref{fig_col_dist_est_env}(d), respectively. For Fig.~\ref{fig_col_dist_est_env}(e), we randomly sampled a configuration of objects and then trained DiffCo on it. Then, when we test the DiffCo, we performed active learning that updates the learned model to adapt to the unforeseen environment. Further, to obtain ground truths, we used FCL as a baseline. 

To compare the performance, we use three metrics (i.e., MAE, AUC, and ACD) in addition to the elapsed time. The MAE represents the mean absolute error between the estimated collision distance and its ground truth from FCL. AUC is the area under the receiver operating characteristic (ROC) curve metric for the binary collision classification problems. The ACD is the average of cosine distances between the estimated gradient fields and their ground truths from FCL (see Fig~\ref{fig:gradients}). To obtain gradient value per environment, we randomly sampled 100 gradients over the space.

\subsection{Trajectory Optimization with Gradients}
We then evaluated the planning performance with our GraphDistNet. Our differential collision detection can be combined with any TO algorithms, such as CHOMP~\cite{zucker2013chomp} and TrajOpt~\cite{schulman2013finding}, that require a collision cost or constraint function as well as its Jacobian.
We defined a distance cost function $C(d)=\frac{c}{||d-d_{margin}||^2+\epsilon}$, where $c$ is an constant and $d$ is the collision distance that is a function of joint angles $\vtheta$. By obtaining the gradient $\partial C(d) /\partial \vtheta$ via the auto differentiation in Pytorch~\cite{NEURIPS2019_9015}, we performed trajectory optimization of $20$ waypoints given a linearly interpolated joint trajectory. In this evaluation, we used an Adam optimizer similar to DiffCo~\cite{zhi2022diffco}. We performed 20 random reaching evaluations per setting, where we used three types of random environments such as Fig.~\ref{fig_col_dist_est_env}(b), Fig.~\ref{fig_col_dist_est_env}(d), and Fig.~\ref{fig:representative}. Note that we perform the iterations until the cost falls below a certain threshold.

To measure the TO performance, we used three metrics: average elapsed time, average path cost, and reaching success rate. The path cost is the sum of distances between end-effector waypoints: $\sum_t ||\vtheta_{t+1}-\vtheta_t||^2$. For a fair comparison, we considered the cost data collected from environments where all evaluation methods succeeded.

\subsection{Demonstration with Manipulation Tasks}
We then demonstrated the feasibility of our method via real-world experiment. We set an object-fetch scenario, in which a robot has to reach and grasp a target object avoiding any environmental collisions. We used a robot that is a 6-DoF manipulator (i.e., UR5e from Universal Robots) with a parallel-jaw gripper (i.e., 2F-85 from Robotiq). To simplify the demonstration task, we used the second, third, and fourth joints (i.e., \textit{shoulder\_lift\_joint}, \textit{elbow\_joint}, \textit{wrist\_1\_joint}) only, otherwise we fixed the remained joints to make the end effector move on a vertical plane. 

For training GraphDistNet and another baseline method, DiffCo, we performed three steps of training processes. 1) We first represented the complex geometry of the manipulator as a graph $G_r$ by projecting the link meshes on the vertical plane and manually extracting the boundaries of them as polygons by using a point-based annotation method (see Fig.~\ref{fig:representative}). We extracted multiple polygons and combined them as a $G_r$ based on joint angles considering the revolute joints. Likewise, we also created $G_o$. %
2) We then prepared a collision-distance dataset with $50,000$ samples that consist of graphs $(G_r, G_o)$ and their distances. For graphs, we sampled joint angles from a uniform random distribution and also varied the gripper polygon considering the size of an object in hand. For distance computation, we used FCL. 
3) Finally, we generated an optimal joint trajectory via TO given initial and final joint angles for fetching. For the optimization, we used a sum of three costs: position error, joint-limit violation, and collision distance. We provided a linearly interpolated joint trajectory as an initial guess of TO.

\section{Evaluation\label{sec:eval}}
\vspace{-4pt}
\begin{figure}[t]
\centering
\includegraphics[width=0.93\columnwidth]{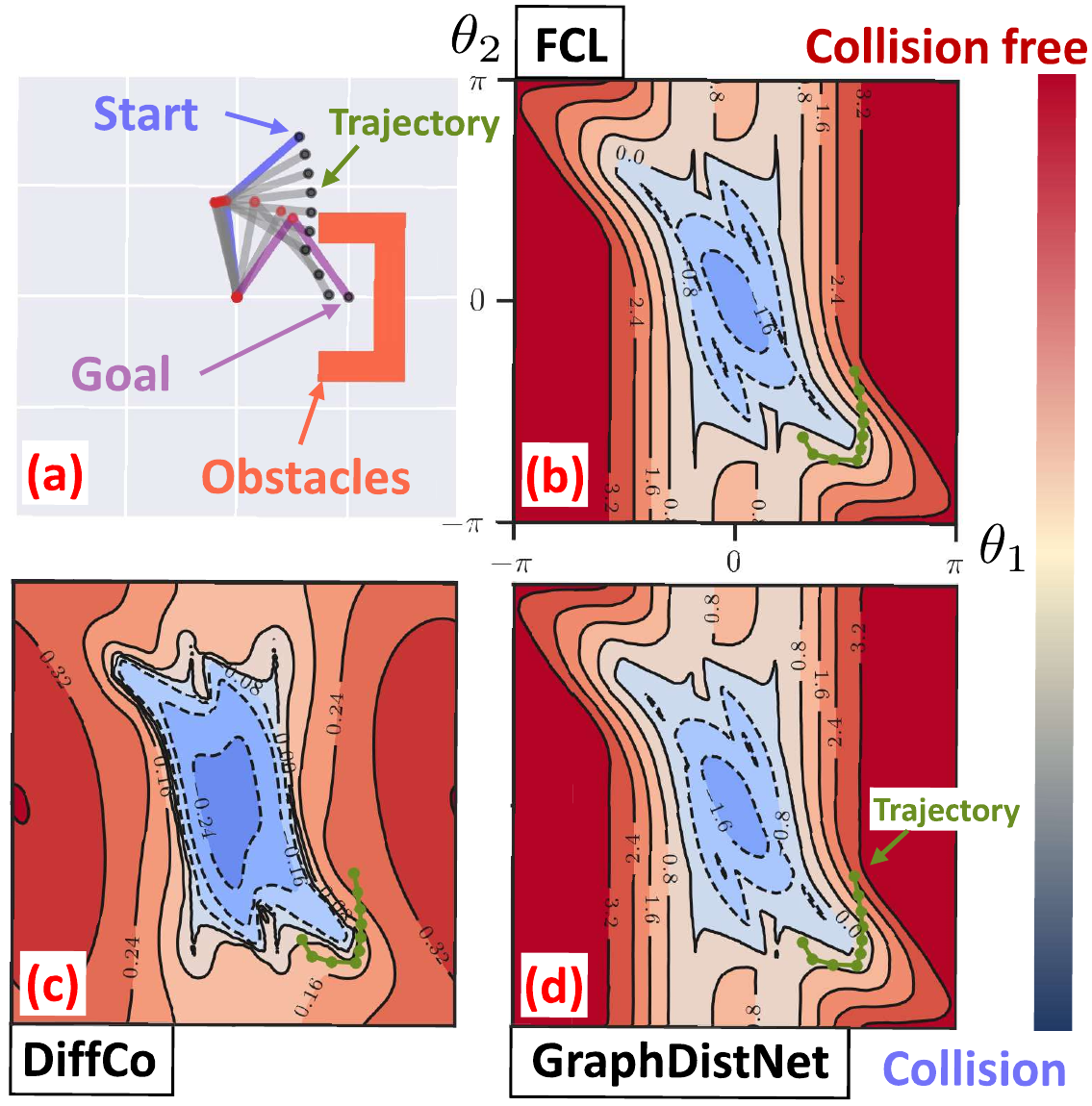}
\vspace{-4pt}
\caption{\small{
Exemplar plots of the minimum distance (or collision score) field given a 2-DoF planar arm and an obstacle. We use FCL, DiffCo, and GraphDistNet to generate each field in order. The contour lines represent the change of distance with respect to the configuration of the arm (i.e., gradient), where the blue and red spaces represent the likelihood of collision and collision-free spaces, respectively. We projected the trail of arms in (a) as an end-effector trajectory (green) in (b), (c), and (d).
}}
\label{fig:gradients}
\vspace{-16pt}
\end{figure}

\begin{figure*}[t]
\centering
\includegraphics[width=2\columnwidth]{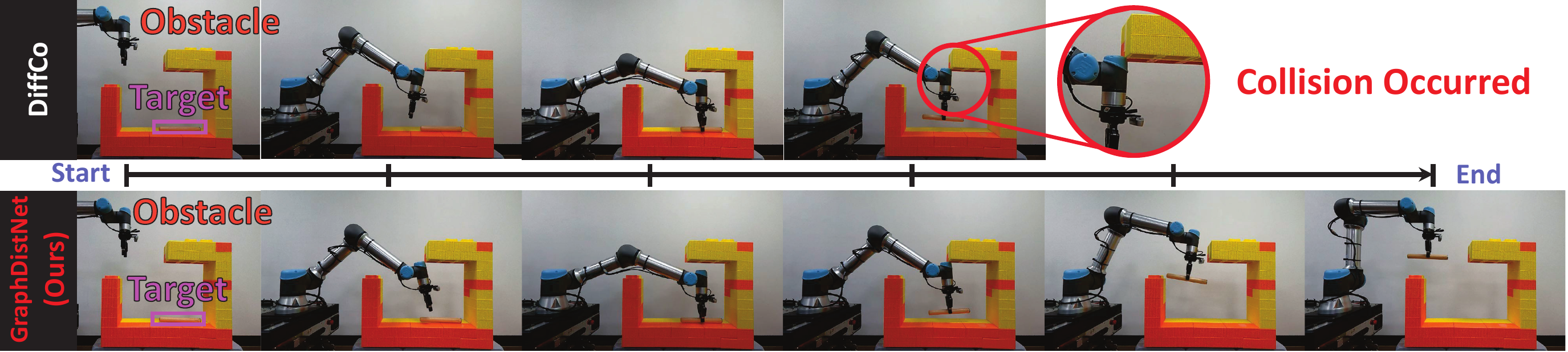} 
\caption{\small{Demonstrations on an object-fetch task. Each row shows a UR5e robot tracking a joint space trajectory computed via trajectory optimization with either DiffCo or GraphDistNet collision checkers. The result of DiffCo shows collisions with the storage (orange).} }
\label{fig:demo}
\vspace{-18pt}
\end{figure*}

We first compared the estimation performance of GraphDistNet with that of three baseline methods through five types of clutter environments. Table~\ref{table_dist_grad_comp} shows our GraphDistNet resulted in the lowest distance and gradient estimation errors (i.e., MAE and ACD) as well as superior collision-detection performances (i.e., AUC) in most environments, except the environment (d). In environment (d), the DiffCo with an active learning scheme could show the lowest ACD value. However, the method required significantly longer adaptation time (i.e., about \SI{15}{\second} per environment) and failed to run in the most complex environment (e). Such disadvantages will largely degrade the usability of potential applications. In addition, another neural network based approach, ClearanceNet, could show the second best performance but ClearanceNet also resulted in large accuracy drops in complex environments such as Fig.~\ref{fig_col_dist_est_env} (c), (d), and (e). 

In this work, we particularly compared the gradient estimation performance toward TO. Fig.~\ref{fig:gradients} shows the exemplar visualizations of gradient fields from FCL, DiffCo, and GraphDistNet. Unlike DiffCo, the field of GraphDistNet is indistinguishable from that of ground truth from FCL. Thus, we can expect that our method is potentially useful for precise optimization by guiding the trajectory (green) to avoid collision precisely following the collision gradients. On the other hand, as shown in Fig.~\ref{fig:gradients} (c), the inaccurate or drastically changing gradients of DiffCo may lead to unstable optimization reaching local minima.

Table~\ref{tab:trajopt} shows the comparison of TO performance given three challenging environments. In simple environments (i.e., Fig.~\ref{fig_col_dist_est_env} (b)), differences between algorithms appeared less though DiffCo was able to return a quick solution. Given more complex environments, our method showed consistently high success rates in finding the best available paths. Although ClearanceNet also resulted in low path costs, its success rate is significantly low. As mentioned before, the fixed representation of the input in ClearanceNet restricts the use in a new complex environment. The restriction also affects the TO performance. Note that our method takes a comparably long time since we did not utilize the batch prediction capability of GraphDistNet. If we adopt batch gradient optimization methods in TO, we expect to show superior efficiency in the future.

\begin{table}[t]
  \centering
  \begin{tabular}{clccc}
     \toprule
     \multirow{2}{*}{Env.}  & \multicolumn{1}{c}{Collision} & Avg. Elapsed & Avg. Path  & Success \\
     & \multicolumn{1}{c}{Checker} & Time (s) & Cost & Rate ($\%$) \\
     \midrule
     \multirow{3}{*}{\shortstack[lb]{Fig.~\ref{fig_col_dist_est_env}\\(b)}}
                & DiffCo & 0.9304 & \textbf{23.7657} & 0.95\\
                & ClearanceNet & 1.7300 & 24.9018 & \textbf{1.0} \\
                & GraphDistNet & 1.9522 & 24.4297 & \textbf{1.0}\\
      \hline
      \multirow{3}{*}{\shortstack[lb]{Fig.~\ref{fig_col_dist_est_env}\\(d)}}
                & DiffCo & 28.8247 & 99.0850 & \textbf{0.95}\\
                & ClearanceNet & 10.9816 & 72.3153 & 0.4 \\
      			& GraphDistNet & 16.2682 & \textbf{65.0297} & 0.9\\
      \hline
            \multirow{3}{*}{\shortstack[lb]{Fig.~\ref{fig:representative}}}
              & DiffCo & 0.6364 & 3.9$\times 10^5$ & 0.4 \\
              & ClearanceNet & 1.9603 & 5.5$\times 10^5$ & 0.3 \\
              & GraphDistNet & 3.2982 & 9.0$\times 10^5$ & \textbf{0.7} \\
     \bottomrule
  \end{tabular}
  \caption{\small{Comparison of trajectory optimization (TO) performance. We performed 20 random evaluations per method that is combined with a TO method. %
  }}
  \label{tab:trajopt}
  \vspace{-24pt}
\end{table}

We finally demonstrated our GraphDistNet-based TO method with a real UR5e manipulator. Fig.~\ref{fig:demo} shows an object fetch scenario, where we generated joint trajectories from DiffCo and GraphDistNet combining a TO approach. Given GraphDistNet, the manipulator could successfully reach a target object (i.e., a wood stick) avoiding collisions, particularly around the small entrance of the storage (orange). Further, our method was able to safely bring out the object in the gripper by training the GraphDistNet with random object geometries. On the other hand, the trajectory generation with DiffCo could precisely reach the target after active learning. However, the trajectory from DiffCo failed to bring the target object back without risk since the simple perceptron model is not enough to estimate the accurate collision scores required for planning in a narrow passage. The limitation of DiffCo significantly degrades the usability of the system, pausing the manipulation process. Overall, the demonstration shows our robot achieves effective trajectory optimization in a real-world manipulation scenario.
Note that we demonstrated a planar TO problem with 2D coordinate features aligning the robot and obstacles on the vertical plane. However, GraphDistNet can be extended to more complex scenarios adopting 3D features or graphs from point clouds.

\section{Conclusion}
\label{sec:conclusion}
We introduced a graph-based collision-distance estimation network, GraphDistNet, that encodes the geometric relation between two graphs and predicts their minimum distance and gradients to optimize a collision-free trajectory in clutter. Our method provides efficient and precise estimation performance and generalization capabilities in unforeseen environments by leveraging an attention mechanism. The statistical evaluation of GraphDistNet shows superior performance compared to other baseline methods in various simulation environments. We also demonstrated the applicability in the real world using a UR5e manipulator.

\bibliographystyle{ieeetr}
\bibliography{root}

\end{document}